\documentclass[10pt,twocolumn,letterpaper]{article}

\usepackage{iccv}
\usepackage{times}
\usepackage{epsfig}
\usepackage{graphicx}
\usepackage{amsmath}
\usepackage{amssymb}

\usepackage{multirow}
\usepackage{color}

\iccvfinalcopy 

\def\iccvPaperID{356} 

\def\etal{{\em et al.}}

\newcommand{\bl}[1]{\textbf{#1}}

\newcommand{\mc}[1]{\mathcal{#1}}

\ificcvfinal\fi
\begin{document}

\title{Selectivity or Invariance: Boundary-aware Salient Object Detection}

\author{Jinming Su$^{1,3}$, Jia Li$^{1,3*}$, Yu Zhang$^{1}$, Changqun Xia$^{3}$ and Yonghong Tian$^{2,3*}$\\
$^1$State Key Laboratory of Virtual Reality Technology and Systems, SCSE, Beihang University\\
$^2$National Engineering Laboratory for Video Technology, School of EE\&CS, Peking University\\
$^3$Peng Cheng Laboratory, Shenzhen, China \ \ \\
{\tt\small \{sujm, jiali\}@buaa.edu.cn, zhangyulb@gmail.com, xiachq@pcl.ac.cn, yhtian@pku.edu.cn}
}
\maketitle
\thispagestyle{empty}

\begin{abstract}
Typically, a salient object detection (SOD) model faces opposite requirements in processing object interiors and boundaries. The features of interiors should be invariant to strong appearance change so as to pop-out the salient object as a whole, while the features of boundaries should be selective to slight appearance change to distinguish salient objects and background. To address this selectivity-invariance dilemma, we propose a novel boundary-aware network with successive dilation for image-based SOD. In this network, the feature selectivity at boundaries is enhanced by incorporating a boundary localization stream, while the feature invariance at interiors is guaranteed with a complex interior perception stream. Moreover, a transition compensation stream is adopted to amend the probable failures in transitional regions between interiors and boundaries. In particular, an integrated successive dilation module is proposed to enhance the feature invariance at interiors and transitional regions. Extensive experiments on six datasets show that the proposed approach outperforms 16 state-of-the-art methods.
\end{abstract}
\let\thefootnote\relax\footnotetext{* Correspondence should be addressed to Jia Li and Yonghong Tian. The code and models are available on http://cvteam.net.}

\section{Introduction}
\begin{figure}[t]
\centering
\includegraphics[width=1\columnwidth,height=4.4cm]{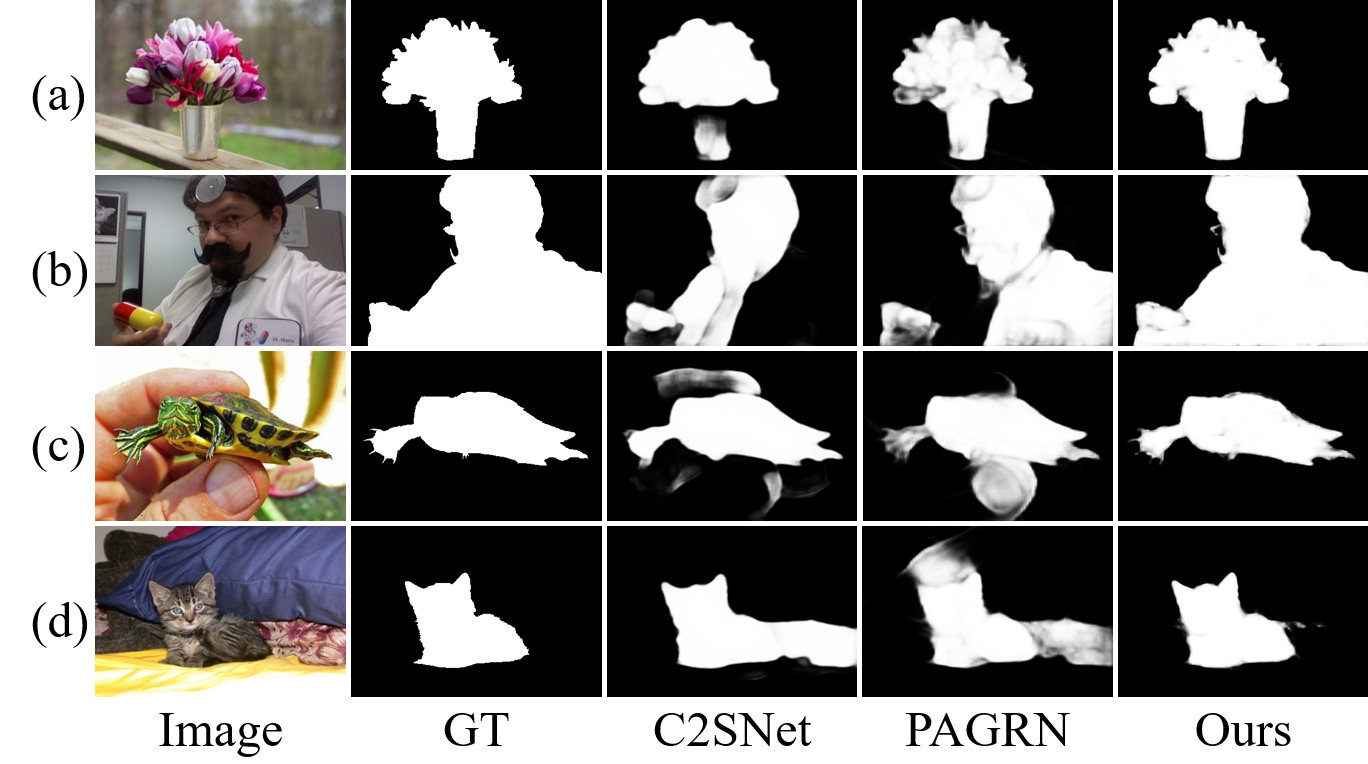}
\caption{Different regions of salient objects require different features. (a)(b)~Features of interiors should be invariant to large appearance change to detect the salient object as a whole; (c)(d) features at boundaries should be selective to distinguish the slight differences between salient objects and background regions. Images and ground-truth masks are from~\cite{yan2013hierarchical}. Results are generated by C2SNet~\cite{li2018contour}, PAGRN~\cite{zhang2018progressive} and our approach. }
\label{fig:motivation}
\end{figure}

Salient object detection (SOD), which aims to detect and segment objects that can capture human visual attention, is an important step before subsequent vision tasks such as object recognition~\cite{ren2014region}, tracking~\cite{hong2015online} and image parsing~\cite{lai2016saliency}. To address the SOD problem, hundreds of learning-based models \cite{jiang2013salient,wang2017stagewise,xia2017and,chen2017look,zhang2017amulet, fan2019shifting} have been proposed in the past decades, among which the state-of-the-art deep models \cite{wang2017stagewise,chen2017look} have demonstrated impressive performance on many datasets~\cite{yan2013hierarchical,yang2013saliency,li2014secrets,
li2015visual,wang2017learning,xia2017and,fan2018salient}. However, there still exist two key issues that need to be further addressed. First, the interiors of a large salient object may have large appearance change, making it difficult to detect the salient object as a whole (see Fig.~\ref{fig:motivation}(a)(b)). Second, the boundaries of salient objects may be very weak so that they cannot be distinguished from the surrounding background regions (see Fig.~\ref{fig:motivation}(c)(d)). Due to these two issues, SOD remains a challenging task even in the deep learning era.

By further investigating these two issues at object interiors and boundaries, we find that the challenge may be mainly from the selectivity-invariance dilemma \cite{lecun2015deep}. In the interiors, the features extracted by a SOD model should be invariant to various appearance changes such as size, color and texture. Such invariant features ensure that the salient object can pop-out as a whole. However, the features at boundaries should be sufficiently selective at the same time so that the minor difference between salient objects and background regions can be well distinguished. In other words, different regions of a salient object poses different requirements for a SOD model, and such dilemma actually prevents the perfect segmentation of salient objects with various sizes, appearances and contexts.


\begin{figure*}[t]
\centering
\includegraphics[width=1\textwidth,height=6.7cm]{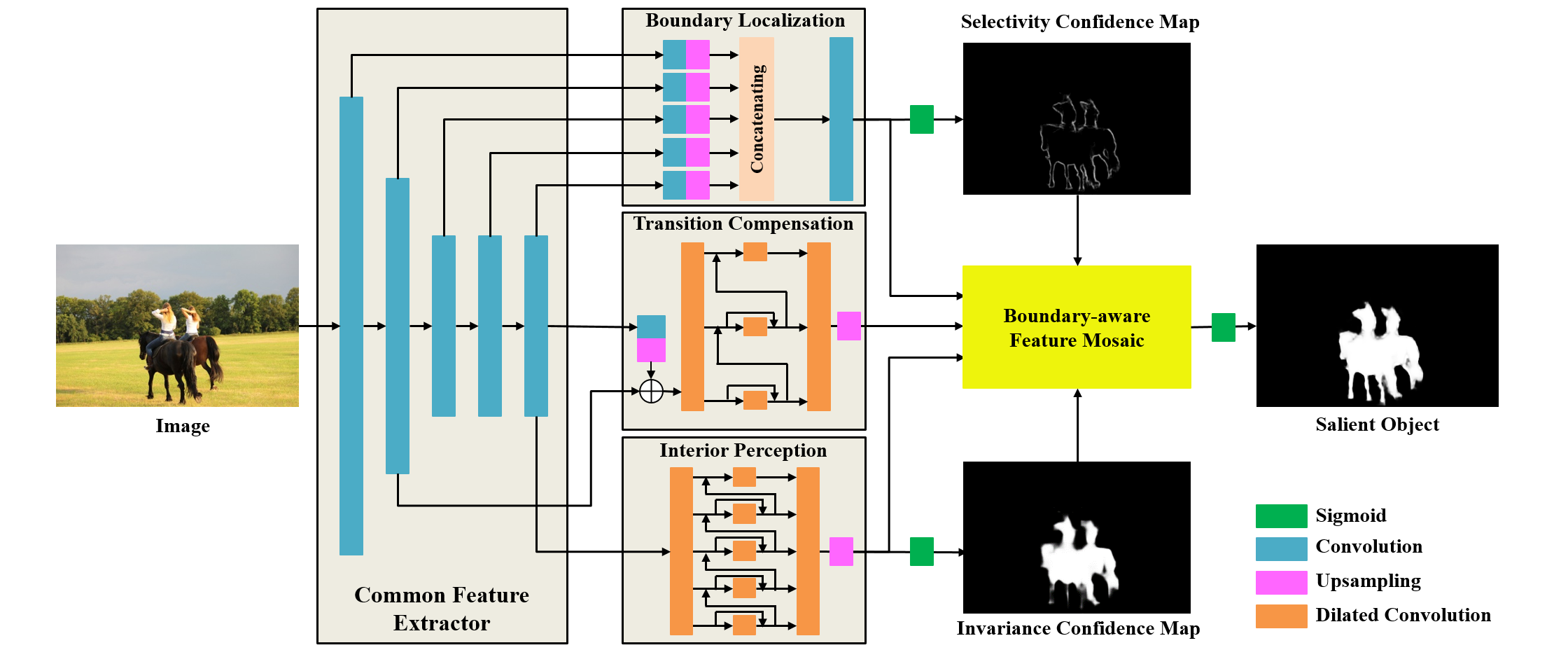}
\caption{The framework of our approach. We first use ResNet-50 to extract common features for three streams. The boundary localization stream uses multi-level features and a simple network to detect salient boundaries with high selectivity, while the interior perception stream uses single-level features and a complex network to guarantee invariance in salient interiors. Their output features are used to form two confidence maps of selectivity and invariance, based on which a transition compensation stream is adopted to amend the probable failures that are likely to occur in the transition regions between boundaries and interiors. These three streams are concatenated to form a boundary-aware feature mosaic map so that the salient object can pop-out as a whole with clear boundaries.}
\label{fig:framework}
\end{figure*}


To break out of this dilemma, a feasible solution is to adopt different feature extraction strategies at object interiors and boundaries. Inspired by that, we propose a boundary-aware network with successive dilation for image-based SOD. As shown in Fig.~\ref{fig:framework}, the network first extracts common visual features and then deliver them into three separate streams. Among these three streams, the boundary localization stream is a simple subnetwork that aims to extract selective features for detecting the boundaries of salient objects, while the interior perception stream emphasizes the feature invariance in detecting the salient objects. In addition, a transition compensation stream is adopted to amend the probable failures that may occur in the transitional regions between interiors and boundaries, where the feature requirement gradually changes from invariance to selectivity. Moreover, an integrated successive dilation module is proposed to enhance the capability of the interior perception and transition compensation streams so that they can extract invariant features for various visual patterns. Finally, the output of these three streams are adaptively fused to generate the masks of salient objects. Experimental results on six public benchmark datasets show that our approach outperforms 16 state-of-the-art SOD models. Moreover, our approach demonstrates impressive capability in accurately segmenting salient boundaries at fine scales.

The main contributions of this paper include: 1)~we revisit the problem of SOD from the perspective of selectivity-invariance dilemma, which may be helpful to develop new models; 2)~we propose a novel boundary-aware network for salient object detection, which consistently outperforms 16 state-of-the-art algorithms on six datasets; 3)~we propose an integrated successive dilation module that can enhance the capability of extracting invariant features.

\section{Related Work}
Hundreds of image-based SOD methods have been proposed in the past decades. Early methods mainly adopted hand-crafted local and global visual features as well as heuristic saliency priors such as the color difference \cite{achanta2009frequency}, distance transformation \cite{tu2016real} and local/global contrast \cite{klein2011center,cheng2015global}. More details about the traditional methods can be found in the survey \cite{borji2015salient}. In this review, we mainly focus on the latest deep models in recent three years.

Lots of these deep models are devoted to fully utilizing the feature integration to enhance the performance of neural networks~\cite{lee2016deep,li2016visual,liu2016dhsnet,wang2017stagewise,zhang2017amulet,luo2017non,zhang2017learning,zhang2018progressive,zhang2018bi}. For example, Zhang~\etal~\cite{zhang2018progressive} proposed an attention guided network to selectively integrates multi-level information in a progressive manner. Wang~\etal~\cite{wang2017stagewise} proposed a pyramid pooling module and a multi-stage refinement mechanism to gather contextual information and stage-wise results, respectively. Zhang~\etal~\cite{zhang2017amulet} adopted a framework to aggregate multi-level convolutional features into multiple resolutions, which were then combined to predict saliency maps in a recursive manner. Luo~\etal~\cite{luo2017non} proposed a simplified convolutional neural network by combining global and local information through a multi-resolution $4 \times 5$ grid structure. Zhang~\etal~\cite{zhang2017learning} utilized the deep uncertain convolutional features and proposed a reformulated dropout after specific convolutional layers to construct an uncertain ensemble of internal feature units. Different with them, we propose an integrated successive dilation module to capture richer contextual information to produce features that account for interior invariance and introduce skip connections from low-level features to promote selective representations of boundaries.

In addition, many models~\cite{wang2016kernelized,chen2017look,chen2018progressively,chen2018eccv,zeng2018learning,wang2018detect,li2018contour} comprehend saliency detection task by relating other vision tasks. Chen \etal~\cite{chen2018eccv} proposed reverse attention mechanism which is inspired from human perception process by using top information to guide bottom-up feed-forward process in a top-down manner. Chen \etal~\cite{chen2017look} incorporated human fixation with semantic information to simulate the human annotation process for salient objects. Chen and Li~\cite{chen2018progressively} proposed a complementarity-aware network to fuse both cross-model and cross-level features to solve saliency detection task with depth information. Wang~\etal~\cite{wang2018detect} proposed to learn the local contextual information for each spatial position to refine boundaries. Li \etal~\cite{li2018contour} considered contours as useful priors and proposed to facilitate feature learning in SOD by transferring knowledge from an existing contour detection model. Our work differs with them by fusing the boundary and interior features of salient objects with a compensation mechanism and an adaptive manner. 


\section{The Proposed Approach}

The selectivity-invariance dilemma in SOD indicates that the boundaries and interiors of salient objects require different types of features. Inspired by that, we propose a boundary-aware network for saliency detection (see Fig.~\ref{fig:framework} for the system framework).
The network first extract common features, which are then processed with three separate streams. The outputs of these streams are then fused to generate the final masks of salient objects in a boundary-aware feature mosaic selection manner.
Details of the proposed approach are descried as follows.

\subsection{Common Feature Extraction}
As shown in Fig.~\ref{fig:framework}, the boundary-aware network starts with ResNet-50~\cite{he2016deep}. As a common feature extractor, we remove the last global pooling and fully connected layers and use only the five residual blocks. For the sake of simplification, the subnetworks in these five blocks are denoted as $\theta_i(\pi_i),i\in\{1,\ldots,5\}$, where $\pi_i$ is the set of parameters of $\theta_i$. Note that the input of $\theta_i(\pi_i)$ is the output of $\theta_{i-1}(\pi_{i-1}),\forall~i=2,\ldots,5$, and we omit the input for the sake of simplification. In addition, the strides of all convolutional layers in $\theta_4$ and $\theta_5$ are set to 1 to avoid the over downsample of feature maps. As in \cite{yu2015multi}, we enlarge the receptive fields by using the dilation of 2 and 4 in all convolutional layers of $\theta_4$ and $\theta_5$, respectively. Finally, for a $H\times{}W$ input image, the subnetwork $\theta_5$ outputs a $\frac{H}{8} \times \frac{W}{8}$ feature map with 2048 channels.

\begin{figure}[t]
\centering
\includegraphics[width=1\columnwidth, height=4cm]{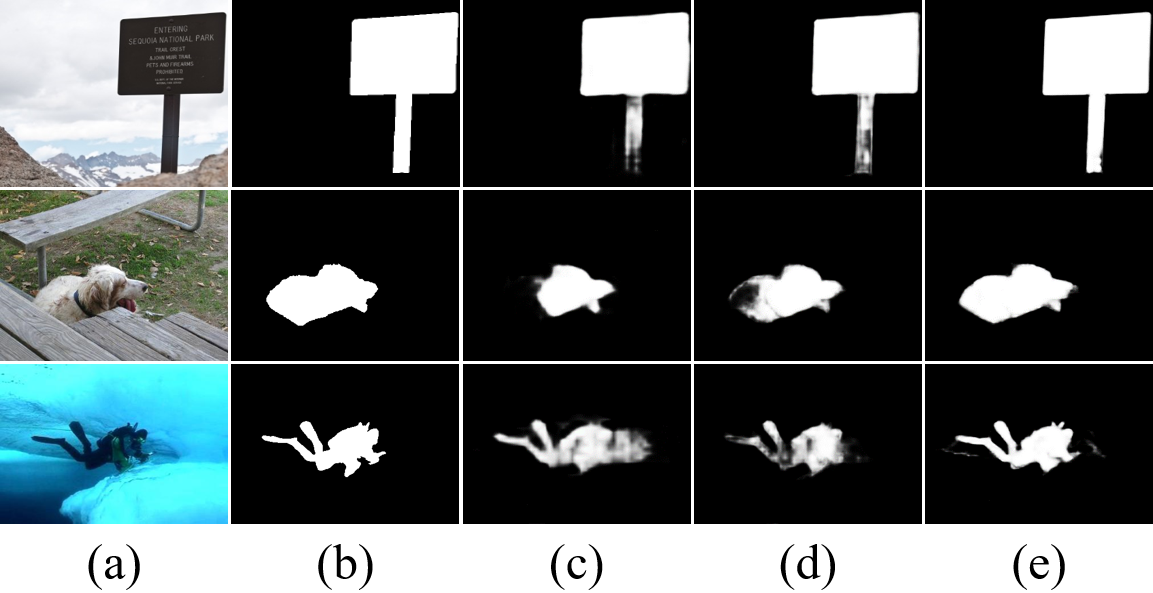}
\caption{The SOD results from the combination of three streams. (a)~Image; (b)~ground-truth; (c)~only interior perception stream; (d)~interior perception and boundary localization streams; (e)~three streams. We can see that interior perception stream may fail near object boundaries due to the emphasis of invariance, while such vague boundaries can be corrected by incorporating the boundary localization stream with the emphasis of selectivity. In addition, the probable failures of these two streams can be amended by the transition compensation stream.}
\label{fig:motivation-of-three-streams}
\end{figure}

\subsection{Boundary-aware SOD with Three Streams}
Given the common features, we use three streams for boundary localization, interior perception and transition compensation, respectively. The boundary stream is inspired by the work of \cite{xie2015holistically}which is a simple subnetwork $\phi_{\mc{B}}(\pi_\mc{B})$ that aggregates multi-level common features and fuses them by upsampling and concatenating to obtain the final boundary predictions. The input of this subnetwork is the concatenation of features from $\{\theta_i(\pi_i)\}_{i=1}^5$. For the feature map of each $\theta_i(\pi_i)$, we add two convolution layers with 128 kernels of $3\times{}3$ and one $1\times{}1$ kernel, respectively. These two layers are used to squeeze the common features, which are then upsampled to $H\times{}W$. After the concatenation, we add an extra layer with one $1\times{}1$ kernel to output a single channel $H\times{}W$ feature map $\phi_{\mc{B}}(\pi_\mc{B})$. A sigmoid layer is then used to generate a selectivity confidence map that is expected to approximate the boundary map of salient objects (denoted as $G_\mc{B}$) by minimizing the loss
\begin{equation}\label{eq:boundaryloss}
L_\mc{B}=E(Sig(\phi_{\mc{B}}(\pi_\mc{B})),G_\mc{B}),
\end{equation}
where $Sig(\cdot)$ is the sigmoid function and $E(\cdot, \cdot)$ means the cross-entropy loss function. By taking multi-level features as the input and using only simple feature mapping subnetworks, the boundary localization stream demonstrates a strong selectivity at object boundaries.

Different from the boundary localization stream, the interior perception stream $\phi_{\mc{I}}(\pi_\mc{I})$ emphasizes feature invariance inside large salient objects. Therefore, it takes less input features and uses a more complex subnetwork. Its input is the output of the last common feature extractor $\theta_5(\pi_5)$, and the output is a single-channel $H\times{}W$ feature map $\phi_{\mc{I}}(\pi_\mc{I})$. Similarly, we can use the sigmoid operation to derive an invariance confidence map, which is expected to approximate the ground-truth mask of salient objects $G$ by minimizing the cross-entropy loss:
\begin{equation}\label{eq:interiorloss}
L_\mc{I}=E(Sig(\phi_{\mc{I}}(\pi_\mc{I})),G),
\end{equation}
Note that an integrated successive dilation (ISD) module is used in this stream to map the input to the output by perceiving local contexts at successive scales, which will be introduced in the next subsection.

As shown in Fig.~\ref{fig:motivation-of-three-streams}, the awareness of boundaries can be enhanced by handling the boundaries and interior regions with two separate streams: one uses multi-level features and a simple network to emphasize selectivity, the other one uses single-level features and a complex network to enhance invariance. However, the combination of these two streams may still have failures, especially for the transitional regions between interiors and boundaries that require a balance of selectivity and invariance. To this end, we adopt a transition compensation stream $\phi_{\mc{T}}(\pi_\mc{T})$ to adaptively amend these failures by compensating features in the transitional regions. Different from the first two streams, $\phi_{\mc{T}}(\pi_\mc{T})$ takes the element-wise summation of the two-level features (one high-level $\theta_5(\pi_5)$ and one low-level $\theta_2(\pi_2)$) as the input. In this manner, localization-aware fine-level features and semantic-aware coarse-level ones can jointly enrich the representation power within transition regions.
Since $\theta_2(\pi_2)$ has the resolution $\frac{H}{4} \times \frac{W}{4}$, we upsample $\theta_5(\pi_5)$ to $\frac{H}{4} \times \frac{W}{4}$ after two pre-processing layers using 256 kernels of $3{}\times{}3$ and 256 kernel of $1\times{}1$, respectively. Based on these features, an ISD module with medium complexity is used to generate a transitional feature representation map that mediates both selectivity and invariance, which ensures that detailed structures of salient objects to be correctly detected.

\begin{figure}[t]
\centering
\includegraphics[width=1\columnwidth,height=4cm]{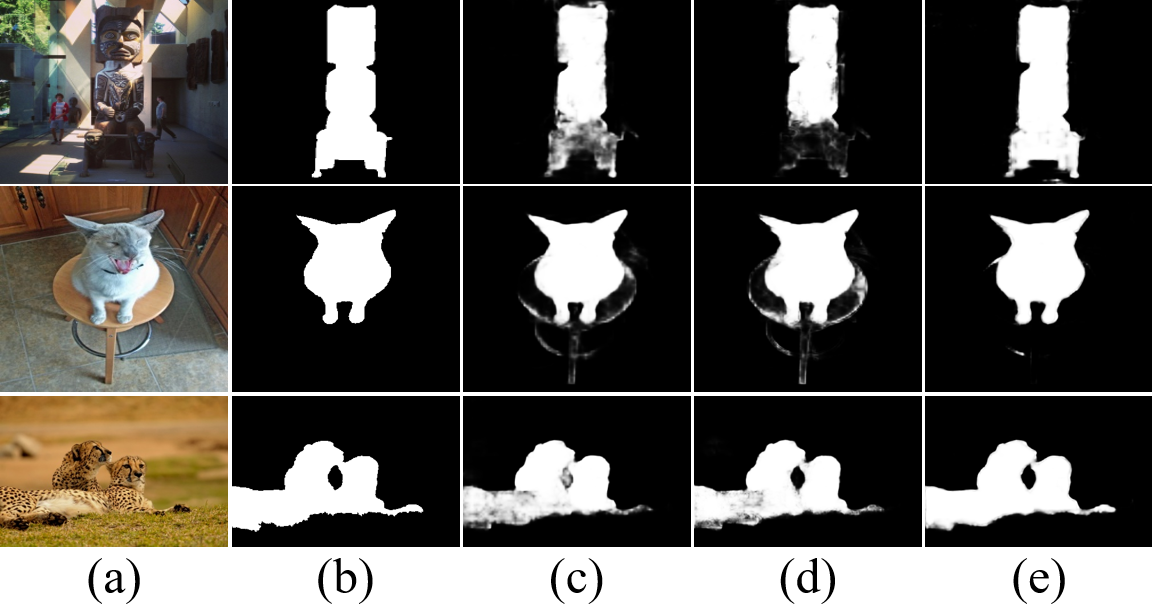}
\caption{Results of different combinations of three streams $\phi_{\mc{B}}$, $\phi_{\mc{I}}$ and $\phi_{\mc{T}}$. (a) Image; (b) ground-truth; (c) element-wise summation; (d) concatenation; (e) our mosaic approach.}
\label{fig:mosaic_motivation}
\end{figure}

Instead of approximating certain ``ground-truth'', the parameters of the transition compensation stream are supervised by the feedback from both the ground-truth masks of salient objects and the predictions of boundary and interior streams. Suppose that this stream also outputs a $H\times{}W$ feature map $\phi_{\mc{T}}$ after upsampling, we combine it with the feature maps $\phi_{\mc{B}}$ and $\phi_{\mc{I}}$. Note that features in $\phi_{\mc{B}}$, $\phi_{\mc{I}}$ and $\phi_{\mc{T}}$ emphasize selectivity, invariance and their tradeoff, respectively. As a result, the direct element-wise summation or concatenation may incorporate unexpected noises as shown in Fig.~\ref{fig:mosaic_motivation}. To reduce these noises, we adopt a boundary-aware mosaic approach that
assigns different strengths to different regions, guided by confidence confidence maps from boundary and interior streams. This approach ensures a well-learned combination of $\phi_{\mc{T}}$ and $\phi_{\mc{B}}$ as well as $\phi_{\mc{I}}$ by properly balancing selectivity and invariance.
Let $\bl{M}$ be the feature mosaic map, we combine the three maps according to the selectivity confidence map $\bl{M}_\mc{B}=Sig(\phi_{\mc{B}})$ and the invariance confidence map $\bl{M}_\mc{I}=Sig(\phi_{\mc{I}})$:
\begin{equation}\label{eq:boundary-aware-feature-fusion}
\begin{split}
\bl{M} = & \phi_{\mc{B}} \otimes (1 - \bl{M}_\mc{I}) \otimes \bl{M}_\mc{B} + \phi_{\mc{I}} \otimes \bl{M}_\mc{I} \otimes (1 - \bl{M}_\mc{B}) \\
            +& \phi_{\mc{T}} \otimes (1 - \bl{M}_\mc{I}) \otimes (1 - \bl{M}_\mc{B}),
\end{split}
\end{equation}
where $\otimes$ denotes the element-wise product between two matrices. We can see that the first term emphasizes selective features $\phi_{\mc{B}}$ at the locations with high selectivity and low invariance confidences, while the second term emphasizes invariant features $\phi_{\mc{I}}$ at the locations with high invariance and low selectivity confidence. For the other locations with medium selectivity and invariance confidences, the transitional features $\phi_{\mc{T}}$ will be added to correct the features. In other words, the transition stream actually learns to approximate the uncertain regions in the other two streams by providing feature compensations. After that, we can derive final saliency map as $Sig(\bl{M})$ by minimizing the loss
\begin{equation}\label{eq:finalLoss}
L_0=E(Sig(\bl{M}),G),
\end{equation}
which indirectly supervises the training process of $\phi_{\mc{T}}(\pi_\mc{T})$. By taking the losses of Eqs.~\eqref{eq:boundaryloss}, \eqref{eq:interiorloss} and \eqref{eq:finalLoss}, the overall learning objective can be formulated as
\begin{equation}\label{eq:overalloss}
\min_{\{\pi_i\}_{i=1}^5,\pi_\mc{B},\pi_\mc{I},\pi_\mc{T}}L_0+L_\mc{B}+L_\mc{I}.
\end{equation}
From Eq.~\eqref{eq:overalloss}, we can see that the parameters $\{\pi_i\}_{i=1}^5$ and $\pi_\mc{T}$ are supervised by the three losses, while the parameters $\pi_\mc{B},\pi_\mc{I}$ are supervised by two losses. Note that the boundary information is used in $L_0$, $L_\mc{B}$, and the generation of the feature mosaic map $\bl{M}$ is also guided by the selectivity confidence map, making the whole network aware of object boundaries.

 \begin{figure}[t]
\centering
\includegraphics[width=1\columnwidth,height=5.5cm]{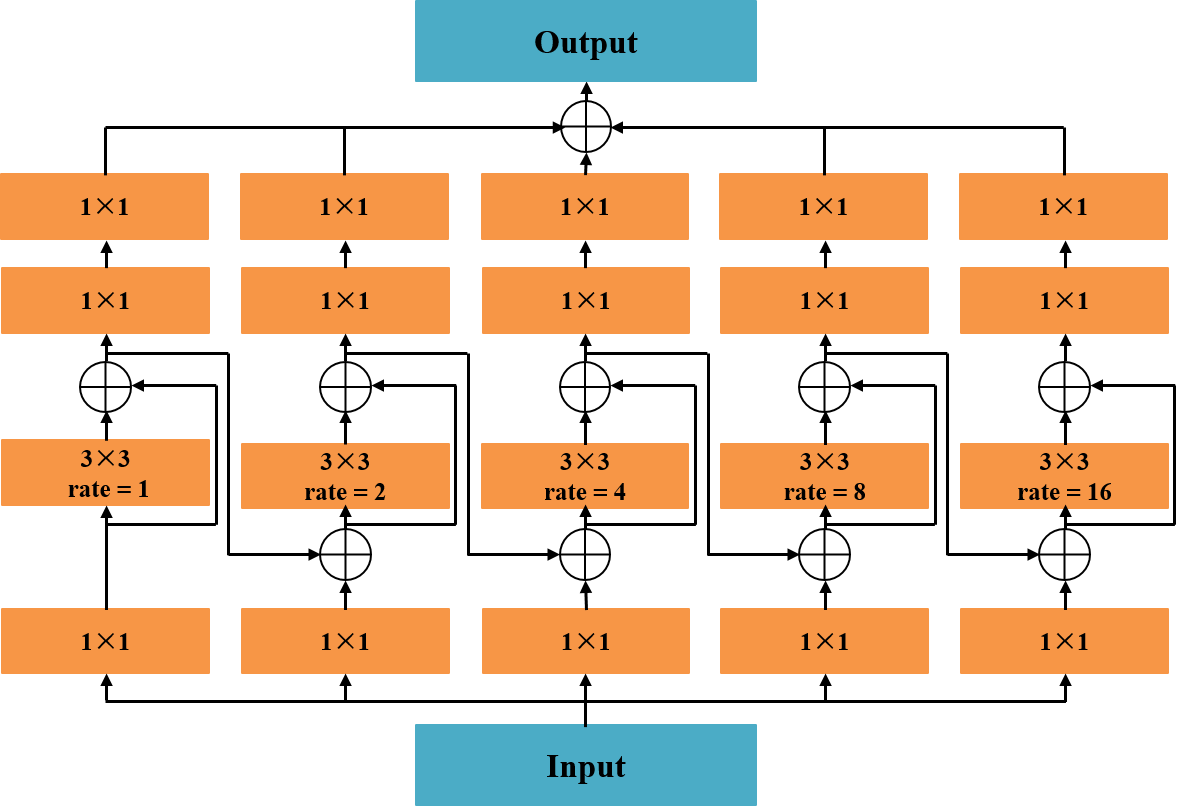}
\caption{Structure of the integrated successive dilation (ISD) module. $1~\times{}1$ and $3~\times{}3$ means the convolutional kernel size, and rate represents the dilation rate in dilated convolution.}
\label{fig:icd}
\end{figure}

\subsection{Integrated Successive Dilation Module}
In the interior perception stream and the transition compensation streams, the key requirement is to extract invariant features for a region embedded in various contexts. To enhance such capability, we propose an integrated successive dilation module (named as ISD) to efficiently aggregate contextual information at a sequence of scales for the purpose of enhancing the feature invariance.

The ISD module with $N$ parallel branches with skip connections is denoted as ISD-$N$, and we show the structure of ISD-5 in Fig.~\ref{fig:icd} as an example. The first layer of each branch is a convolutional layer with $1\times{}1$ kernels that is used for channel compression. The second layer of each branch adopts dilated convolution, in which the dilation rates start from 1 in the first branch and double in the subsequent branch. In this manner, the last branch has a dilation rate of $2^{N-1}$. By adding intra- and inter-branch short connections, the feature map generated by a branch layer actually integrates the perception results of the previous branch and further handle them with larger dilation. In this way, the feature map from the first branch of the second layer is also encoded in the feature maps of subsequent branches, which actually gets processed by successive dilation rates. In other words, an ISD-$N$ module gains the capability of perceiving various local contexts with the smallest dilation rate of 1 and the largest dilation rate of $2^N-1$. After that, the third and the forth layers adopt $1\times{}1$ kernels to integrate feature maps formed under various dilation rates. In practice, we use ISD-5 in the interior perception stream and ISD-3 in the transition compensation streams.

\section{Experiments and Results}
\subsection{Experimental Setup}
\textbf{Datasets.}
To evaluate the performance of the proposed approach, we conduct experiments on six benchmark datasets~\cite{yan2013hierarchical,yang2013saliency,li2014secrets,li2015visual,wang2017learning,xia2017and}. Details of these datasets are described briefly as follows: ECSSD~\cite{yan2013hierarchical} contains 1,000 images with complex structures and obvious semantically meaningful objects. DUT-OMRON~\cite{yang2013saliency} consists of 5,168 complex images with pixel-wise annotations of salient objects and all images are downsampled to a maximal side length of 400 pixels. PASCAL-S~\cite{li2014secrets} includes 850 natural images that are pre-segmented into objects or regions and free-viewed by 8 subjects in eye-tracking tests for salient object annotation. HKU-IS~\cite{li2015visual} comprises 4,447 images and lots of images contain multiple disconnected salient objects or salient objects that touch image boundaries. DUTS~\cite{wang2017learning} is a large scale dataset containing 10533 training images (denoted as DUTS-TR) and 5019 test images(denoted as DUTS-TE). The images are challenging with salient objects that occupy various locations and scales as well as complex background. XPIE~\cite{xia2017and} has 10000 images covering a variety of simple and complex scenes with salient objects of different numbers, sizes and positions.

\textbf{Evaluation Metrics.}
We adopt mean absolute error (MAE), F-measure (F$_{\beta}$) score, weighted F-measure (F$^w_{\beta}$) score \cite{margolin2014evaluate}, Precision-Recall (PR) curve and F-measure curve as our evaluation metrics. MAE reflects the average pixel-wise absolute difference between the estimated and ground-truth saliency maps. In computing F$_{\beta}$, we normalize the predicted saliency maps into the range of [0, 255] and binarize the saliency maps with a threshold sliding from 0 to 255 to compare the binary maps with ground-truth maps. At each threshold, Precision and Recall can be computed. F$_{\beta}$ is computed as:
\begin{equation}\label{eq:Fbeta}
F_{\beta} = \frac{(1 + \beta^2) \cdot Precision \cdot Recall}{\beta^2 \cdot Precision + Recall}.
\end{equation}
where we set $\beta^2 = 0.3$ to emphasize more on Precision than Recall as suggested in \cite{achanta2009frequency}. Then we can plot the PR curve and F-measure curve based on all the binary maps over all saliency maps in a given dataset.

We report F$_{\beta}$ using an adaptive threshold for generating a binary saliency map and the threshold is computed as twice the mean of a saliency map. Besides, F$^w_{\beta}$ is computed to reflect the overall performance (refer to \cite{margolin2014evaluate} for details).

\begin{table*}[t]
\centering
\caption{Performance of 16 state-of-the-arts and the proposed method on six benchmark datasets. Smaller MAE, larger F$^{w}_{\beta}$ and F$_{\beta}$ correspond to better performance. The best results of different backbones are in \textbf{\color{blue}{blue}} and \textbf{\color{red}{red}} fonts. ``-" means the results cannot be obtained and ``$^\dagger$'' means the results are post-processed by dense conditional random field (CRF)~\cite{krahenbuhl2011efficient}. Note that the backbone of PAGRN is VGG-19~\cite{simonyan2015very} and the one of R3Net is ResNeXt-101~\cite{xie2017aggregated}. 
MK: MSRA10K~\cite{cheng2015global}, DUTS: DUTS-TR~\cite{wang2017learning}, MB: MSRA-B~\cite{liu2010learning}.}
\setlength{\tabcolsep}{0.5mm}{
\renewcommand\arraystretch{1.0}
\small
\begin{tabular}{|c|c|c|c|c|c|c|c|c|c|c|c|c|c|c|c|c|c|c|c|}
\hline
\multirow{2}*{Models} & Training & \multicolumn{3}{|c|}{ECSSD} & \multicolumn{3}{|c|}{DUT-OMRON} & \multicolumn{3}{|c|}{PASCAL-S} & \multicolumn{3}{|c|}{HKU-IS} & \multicolumn{3}{|c|}{DUTS-TE} & \multicolumn{3}{|c|}{XPIE}\\
\cline{3-20}
& dataset & MAE & F$^{w}_{\beta}$ & F$_{\beta}$ & MAE & F$^{w}_{\beta}$ & F$_{\beta}$ & MAE & F$^{w}_{\beta}$ & F$_{\beta}$
& MAE & F$^{w}_{\beta}$ & F$_{\beta}$ & MAE & F$^{w}_{\beta}$ & F$_{\beta}$ & MAE & F$^{w}_{\beta}$ & F$_{\beta}$ \\
\hline
\multicolumn{20}{|c|}{VGG-16 backbone~\cite{simonyan2015very}} \\
\hline
KSR~\cite{wang2016kernelized} & MB
            & 0.132 & 0.633 & 0.810 & 0.131 & 0.486 & 0.625 & 0.157 & 0.569 & 0.773 & 0.120 & 0.586 & 0.773 & - & - & - & - & - & - \\
HDHF~\cite{li2016visual} & MB
            & 0.105 & 0.705 & 0.834 & 0.092 & 0.565 & 0.681 & 0.147 & 0.586 & 0.761 & 0.129 & 0.564 & 0.812 & - & - & - & - & - & - \\
ELD~\cite{lee2016deep} & MK
            & 0.078 & 0.786 & 0.829 & 0.091 & 0.596 & 0.636 & 0.124 & 0.669 & 0.746 & 0.063 & 0.780 & 0.827 & 0.092 & 0.608 & 0.647 & 0.085 & 0.698 & 0.746 \\
UCF~\cite{zhang2017learning} & MK
            & 0.069 & 0.807 & 0.865 & 0.120 & 0.574 & 0.649 & 0.116 & 0.696 & 0.776 & 0.062 & 0.779 & 0.838 & 0.112 & 0.596 & 0.670 & 0.095 & 0.693 & 0.773 \\
NLDF~\cite{luo2017non} & MB
            & 0.063 & 0.839 & 0.892 & 0.080 & 0.634 & 0.715 & 0.101 & 0.737 & 0.806 & 0.048 & 0.838 & 0.884 & 0.065 & 0.710 & 0.762 & 0.068 & 0.762 & 0.825 \\
Amulet~\cite{zhang2017amulet} & MK
            & 0.059 & 0.840 & 0.882 & 0.098 & 0.626 & 0.673 & 0.099 & 0.736 & 0.795 & 0.051 & 0.817 & 0.853 & 0.085 & 0.658 & 0.705 & 0.074 & 0.743 & 0.796 \\
FSN~\cite{chen2017look} & MK
            & 0.053 & 0.862 & 0.889 & 0.066 & 0.694 & 0.733 & 0.095 & 0.751 & 0.804 & 0.044 & 0.845 & 0.869 & 0.069 & 0.692 & 0.728 & 0.066 & 0.762 & 0.812 \\
C2SNet~\cite{li2018contour} & MK
            & 0.057 & 0.844 & 0.878 & 0.079 & 0.643 & 0.693 & 0.086 & 0.764 & 0.805 & 0.050 & 0.823 & 0.854 & 0.065 & 0.705 & 0.740 & 0.066 & 0.764 & 0.807 \\
RA~\cite{chen2018eccv} & MB
            & 0.056 & 0.857 & 0.901 & 0.062 & 0.695 & 0.736 & 0.105 & 0.734 & 0.811 & 0.045 & 0.843 & 0.881 & 0.059 & 0.740 & 0.772 & 0.067 & 0.776 & 0.836 \\
Picanet~\cite{liu2018picanet} & DUTS
			& 0.046 & 0.865 & 0.899 & 0.068 & 0.691 & 0.730 & \textbf{\color{blue}{0.079}} & 0.775 & 0.821 & 0.042 & 0.847 & 0.878 & 0.054 & 0.747 & 0.770 & 0.053 & 0.799 & 0.841 \\
PAGRN~\cite{zhang2018progressive} & DUTS
            & 0.061 & 0.834 & 0.912 & 0.071 & 0.622 & 0.740 & 0.094 & 0.733 & 0.831 & 0.048 & 0.820 & 0.896 & 0.055 & 0.724 & 0.804 & - & - & -\\
RFCN~\cite{wang2019salient} & MK
            & 0.067 & 0.824 & 0.883 & 0.077 & 0.635 & 0.700 & 0.106 & 0.720 & 0.802 & 0.055 & 0.803 & 0.864 & 0.074 & 0.663 & 0.731 & 0.073 & 0.736 & 0.809 \\
DSS$^\dagger$~\cite{hou2019deeply} & MB
            & 0.052 & 0.872 & \textbf{\color{blue}{0.918}} & 0.063 & 0.697 & \textbf{\color{blue}{0.775}} & 0.098 & 0.756 & 0.833 & 0.040 & 0.867 & \textbf{\color{blue}{0.904}} & 0.056 & 0.755 & \textbf{\color{blue}{0.810}} & 0.065 & 0.784 & 0.849 \\
\hline 
\textbf{BANet} & MK 
			& 0.046 & 0.873 & 0.907 & 0.062 & 0.705 & 0.742 & 0.082 & 0.780 & 0.832 & 0.041 & 0.851 & 0.883 & 0.048 & 0.766 & 0.791 & 0.052 & 0.808 & 0.853 \\
\textbf{BANet} & DUTS 
			& \textbf{\color{blue}{0.041}} & \textbf{\color{blue}{0.890}} & 0.917 & \textbf{\color{blue}{0.061}} & \textbf{\color{blue}{0.719}} & 0.750 & \textbf{\color{blue}{0.079}} & \textbf{\color{blue}{0.794}} & \textbf{\color{blue}{0.839}} & \textbf{\color{blue}{0.037}} & \textbf{\color{blue}{0.869}} & 0.893 & \textbf{\color{blue}{0.046}} & \textbf{\color{blue}{0.781}} & 0.805 & \textbf{\color{blue}{0.048}} & \textbf{\color{blue}{0.822}} & \textbf{\color{blue}{0.862}} \\
\hline  
\multicolumn{20}{|c|}{ResNet-50 backbone~\cite{he2016deep}} \\
\hline  
SRM~\cite{wang2017stagewise} & DUTS
            & 0.054 & 0.853 & 0.902 & 0.069 & 0.658 & 0.727 & 0.086 & 0.759 & 0.820 & 0.046 & 0.835 & 0.882 & 0.059 & 0.722 & 0.771 & 0.057 & 0.783 & 0.841 \\
Picanet~\cite{liu2018picanet} & DUTS
            & 0.047 & 0.866 & 0.902 & 0.065 & 0.695 & 0.736 & 0.077 & 0.778 & 0.826 & 0.043 & 0.840 & 0.878 & 0.051 & 0.755 & 0.778 & 0.052 & 0.799 & 0.843 \\
DGRL~\cite{wang2018detect} & DUTS
            & 0.043 & 0.883 & 0.910 & 0.063 & 0.697 & 0.730 & 0.076 & 0.788 & 0.826 & 0.037 & 0.865 & 0.888 & 0.051 & 0.760 & 0.781 & 0.048 & 0.818 & 0.859 \\
R3$^\dagger$~\cite{deng2018r3net} & MK
            & 0.040 & 0.902 & 0.924 & 0.063 & 0.728 & \textbf{\color{red}{0.768}} & 0.095 & 0.760 & 0.834 & 0.036 & 0.877 & 0.902 & 0.057 & 0.765 & 0.805 & 0.058 & 0.805 & 0.854 \\
\hline 
\textbf{BANet} & DUTS
            & \textbf{\color{red}{0.035}} & \textbf{\color{red}{0.908}} & \textbf{\color{red}{0.929}} & \textbf{\color{red}{0.059}} & \textbf{\color{red}{0.736}} & 0.763 & \textbf{\color{red}{0.072}} & \textbf{\color{red}{0.810}} & \textbf{\color{red}{0.849}} & \textbf{\color{red}{0.032}} & \textbf{\color{red}{0.886}} & \textbf{\color{red}{0.905}} & \textbf{\color{red}{0.040}} & \textbf{\color{red}{0.811}} & \textbf{\color{red}{0.829}} & \textbf{\color{red}{0.044}} & \textbf{\color{red}{0.839}} & \textbf{\color{red}{0.873}} \\
\hline
\end{tabular}}
\label{tab:result-all}
\end{table*}

\begin{figure*}[t]
\centering
\includegraphics[width=1\textwidth,height=4.1cm]{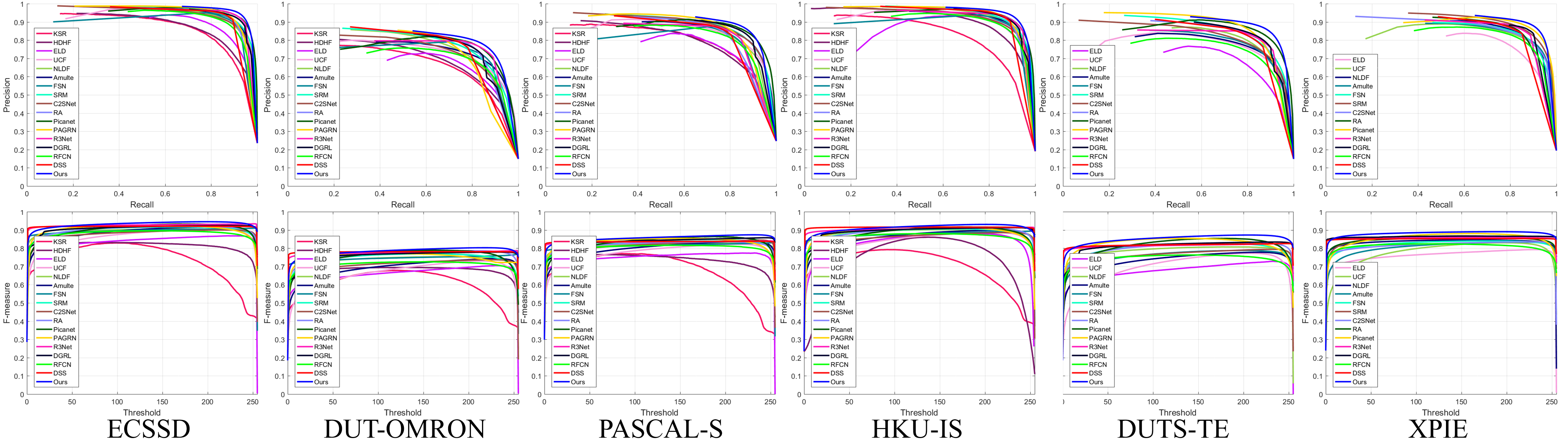}
\caption{The PR curves and F-measure curves of 16 state-of-the-arts and our approach are listed across six benchmark datasets.}
\label{fig:prfmeasure}
\end{figure*}

\begin{figure*}[t]
\centering
\includegraphics[width=1\textwidth,height=8.4cm]{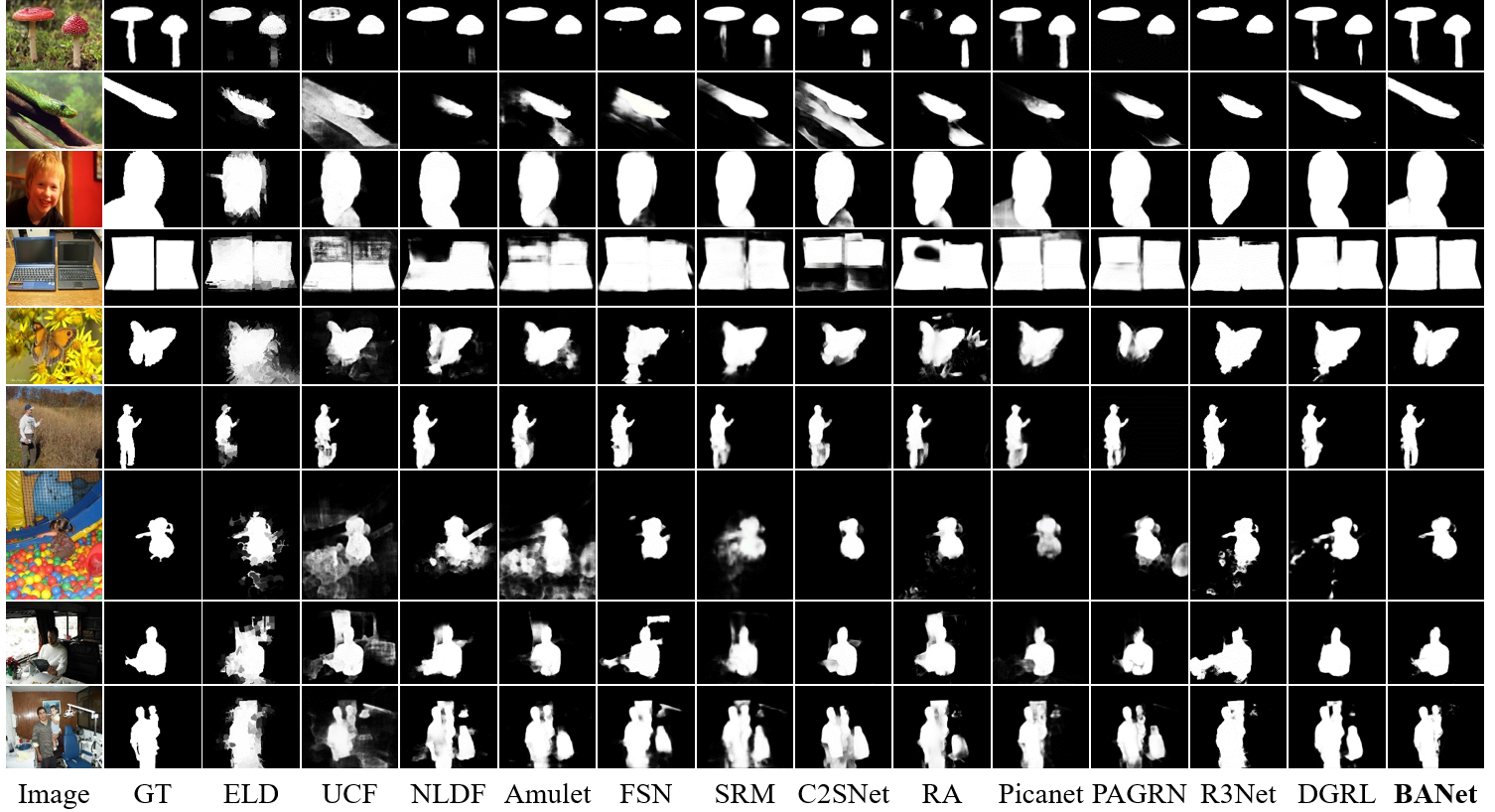}
\caption{Qualitative comparisons of the state-of-the-art algorithms and our approach. GT means ground-truth masks of salient objects. The images are selected from six datasets for testing.}
\label{fig:result-all}
\end{figure*}

\textbf{Training and Inference.}
We use standard stochastic gradient descent algorithm to train the whole network end-to-end with the cross-entropy losses between estimated and ground-truth maps. In the optimization process, the parameter of common feature extractor is initialized by the pre-trained ResNet-50 model \cite{he2016deep}, whose learning rate is set to $5 \times 10^{-9}$ with a weight decay of 0.0005 and momentum of 0.9. The learning rates of the rest layers in our network are set to 10 times larger. Besides, we employ the ``poly'' learning rate policy for all experiments similar to \cite{liu2015parsenet}.

We train our network on DUTS-TR \cite{wang2017learning} as used in~\cite{wang2018detect,liu2018picanet,wang2017stagewise}. For a more comprehensive demonstration, we also trained our network with VGG-16~\cite{simonyan2015very} on MSRA10K~\cite{cheng2015global} as used in~\cite{zhang2017learning,zhang2017amulet,chen2017look,li2018contour} and on DUTS-TR as done in ~\cite{zhang2018progressive,liu2018picanet}. The training images are not done with any special treatment except the horizontal flipping. The training process takes about 15 hours and converges after 200k iterations with mini-batch of size 1. During testing, the proposed network removes all the losses, and each image is directly fed into the network to obtain its saliency map without any pre-processing. The proposed method runs at about 13 fps with about $400 \times 300$ resolution on our computer with a 3.60GHz CPU and a GTX 1080ti GPU.

\subsection{Comparisons with the State-of-the-Arts}
We compare our approach denoted as \textbf{BANet} with 16 state-of-the-art methods, including KSR \cite{wang2016kernelized}, HDHF \cite{li2016visual}, ELD \cite{lee2016deep}, UCF \cite{zhang2017learning}, NLDF \cite{luo2017non}, Amulet \cite{zhang2017amulet}, FSN \cite{chen2017look}, SRM \cite{wang2017stagewise}, C2SNet~\cite{li2018contour}, RA~\cite{chen2018eccv}, Picanet~\cite{liu2018picanet}, PAGRN \cite{zhang2018progressive}, R3Net~\cite{deng2018r3net}, DGRL~\cite{wang2018detect}, RFCN~\cite{wang2019salient} and DSS~\cite{hou2019deeply}. For fair comparison, we obtain the saliency maps of these methods from authors or the deployment codes provided by authors.

\textbf{Quantitative Evaluation.}
The proposed approach is compared with 16 state-of-the-art saliency detection methods on six datasets. The comparison results are shown in Tab.\ref{tab:result-all} and Fig.~\ref{fig:prfmeasure}. From Tab.\ref{tab:result-all}, we can see that our method consistently outperforms other methods across all the six benchmark datasets. It is worth noting that F$^w_{\beta}$ of our method is significantly better compared with the second best results on PASCAL-S (0.810 against 0.788), DUTS-TE (0.811 against 0.765) and XPIE (0.839 against 0.818), and have similar improvements on the other datasets. F$_{\beta}$ also has obvious improvement on all the datasets except DUT-OMRON, on which we achieve the third but the best two methods both employ dense CRF~\cite{krahenbuhl2011efficient} to further refine their results. As for MAE, our approach also achieves the best performance on all the datasets. For overall comparisons, PR and F-measure curve of different methods are displayed in Fig.~\ref{fig:prfmeasure}. One can observe that our approach noticeably outperforms all the other methods. These observations demonstrate the efficiency of boundary-aware network, which indicates that it is useful to deal with the problem of SOD from the perspective of selectivity-invariance dilemma. Note that the results of DSS, RA and HDHF on HKU-IS~\cite{li2015visual} are only conducted on the test set.

\textbf{Qualitative Evaluation.}
Fig.~\ref{fig:result-all} show examples of saliency maps generated by our approach as well as other state-of-the-art methods. We can see that salient objects can pop-out as a whole with clear boundaries by the proposed method. From Fig.~\ref{fig:result-all}, we can find that many methods fail to detect the salient objects with large changed appearance as a whole as depicted in the row of 1 to 3. This indicates the feature invariance is important for SOD, which can be extracted by ISD to guarantee the integrity of salient objects. In addition, when salient objects share the same attributes (such as color, texture and locations) with background, the boundaries of salient objects predicted by many methods become vague, as shown in the row of 4 to 6. In our approach, the feature selectivity at boundary is guaranteed by the awareness of boundaries, which deal with the above situation to obtain clear boundaries. Moreover, three extra examples about more difficult scenes are shown in the last three rows of Fig. \ref{fig:result-all}, our methods also obtain the impressive results. These observations indicated the feature selectivity and invariance are important to deal with the integrity of objects and clarity of boundaries for SOD.

\subsection{Ablation Analysis}
\begin{table*}[t]
\centering
\caption{Performance of the three streams in the proposed approach on six benchmark datasets. IP means interior perception stream, ``IP + BL'' means the combination of interior perception and boundary localization streams, and BANet is our approach.}
\setlength{\tabcolsep}{0.85mm}{
\renewcommand\arraystretch{0.9}
\small
\begin{tabular}{|c|c|c|c|c|c|c|c|c|c|c|c|c|c|c|c|c|c|c|}
\hline
\multirow{2}*{Models} & \multicolumn{3}{|c|}{ECSSD} & \multicolumn{3}{|c|}{DUT-OMRON} & \multicolumn{3}{|c|}{PASCAL-S} & \multicolumn{3}{|c|}{HKU-IS} & \multicolumn{3}{|c|}{DUTS-TE} & \multicolumn{3}{|c|}{XPIE}\\
\cline{2-19}
& MAE & F$^{w}_{\beta}$ & F$_{\beta}$ & MAE & F$^{w}_{\beta}$ & F$_{\beta}$ & MAE & F$^{w}_{\beta}$ & F$_{\beta}$
& MAE & F$^{w}_{\beta}$ & F$_{\beta}$ & MAE & F$^{w}_{\beta}$ & F$_{\beta}$ & MAE & F$^{w}_{\beta}$ & F$_{\beta}$ \\

\hline
IPS
            & 0.048 & 0.868 & 0.902 & 0.068 & 0.679 & 0.723 & 0.084 & 0.774 & 0.823 & 0.043 & 0.839 & 0.873 & 0.050 & 0.753 & 0.779 & 0.052 & 0.801 &0.845 \\
IPS + BLS
            & 0.046 & 0.877 & 0.914 & 0.060 & 0.699 & 0.752 & 0.080 & 0.791 & 0.839 & 0.042 & 0.845 & 0.878 & 0.047 & 0.762 & 0.809 & 0.050 & 0.812 & 0.859 \\
\textbf{BANet}
            & \textbf{0.035} & \textbf{0.908} & \textbf{0.929} & \textbf{0.059} & \textbf{0.736} & \textbf{0.763} & \textbf{0.072} & \textbf{0.810} & \textbf{0.849} & \textbf{0.032} & \textbf{0.886} & \textbf{0.905} & \textbf{0.040} & \textbf{0.811} & \textbf{0.829} & \textbf{0.044} & \textbf{0.839} & \textbf{0.873} \\
\hline
\end{tabular}}
\label{tab:comparisons-of-three-streams}
\end{table*}

\begin{table*}[t]
\centering
\caption{Comparisons of different contextual integration modules on six datasets. ``w/o ISD'' represents BANet without ISD, ``r/w ASPP'' means ISD is replaced with ASPP, and ``r/w ASPP'' means ISD is replaced with ASPP-M.}
\setlength{\tabcolsep}{0.78mm}{
\renewcommand\arraystretch{0.9}
\small
\begin{tabular}{|c|c|c|c|c|c|c|c|c|c|c|c|c|c|c|c|c|c|c|}
\hline
\multirow{2}*{Models} & \multicolumn{3}{|c|}{ECSSD} & \multicolumn{3}{|c|}{DUT-OMRON} & \multicolumn{3}{|c|}{PASCAL-S} & \multicolumn{3}{|c|}{HKU-IS} & \multicolumn{3}{|c|}{DUTS-TE} & \multicolumn{3}{|c|}{XPIE}\\
\cline{2-19}
& MAE & F$^{w}_{\beta}$ & F$_{\beta}$ & MAE & F$^{w}_{\beta}$ & F$_{\beta}$ & MAE & F$^{w}_{\beta}$ & F$_{\beta}$
& MAE & F$^{w}_{\beta}$ & F$_{\beta}$ & MAE & F$^{w}_{\beta}$ & F$_{\beta}$ & MAE & F$^{w}_{\beta}$ & F$_{\beta}$ \\

\hline
w/o ISD
            & 0.046 & 0.876 & 0.913 & 0.064 &0.701 & 0.745 & 0.086 & 0.772 & 0.827 & 0.039 & 0.858 & 0.890 & 0.049 & 0.766 & 0.797 & 0.052 & 0.806 & 0.852 \\
r/w ASPP
            & 0.040 & 0.889 & 0.912 & 0.060 & 0.713 & 0.740 & 0.077 & 0.790 & 0.832 & 0.036 & 0.868 & 0.887 & 0.045 & 0.780 & 0.793 & 0.471 & 0.821 & 0.855 \\
r/w ASPP-M
            & 0.039 & 0.891 & 0.917 & 0.060 & 0.711 & 0.742 & 0.078 & 0.789 & 0.834 & 0.036 & 0.870 & 0.891 & 0.044 & 0.786 & 0.800 & 0.048 & 0.822 & 0.857 \\
\textbf{BANet}
            & \textbf{0.035} & \textbf{0.908} & \textbf{0.929} & \textbf{0.059} & \textbf{0.736} & \textbf{0.763} & \textbf{0.072} & \textbf{0.810} & \textbf{0.849} & \textbf{0.032} & \textbf{0.886} & \textbf{0.905} & \textbf{0.040} & \textbf{0.811} & \textbf{0.829} & \textbf{0.044} & \textbf{0.839} & \textbf{0.873} \\
\hline
\end{tabular}}
\label{tab:comparisons-of-different-modules}
\end{table*}
To validate the effectiveness of different components of the proposed method, we conduct several experiments on all the six datasets to compare the performance variations of our methods with different experimental settings.

\textbf{Effectiveness of the BLS and TCS.} To investigate the efficacy of the proposed boundary localization stream (BLS) and transition compensation stream (TCS), we conduct ablation experiments across all six datasets by introducing two different settings for comparisons. The first setting denoted as ``IPS'' contains only the interior perception stream following the common feature extractor to directly predict the saliency maps. To explore the effectiveness of boundary localization stream, the second one named as ``IPS + BLS'' utilizes the interior perception stream and boundary localization stream together, where the final predicted results are added up directly without the transition compensation stream. Note that our proposed approach \textbf{BANet} combines all three streams.

For a comprehensive comparison, above-mentioned settings and BANet are evaluated on six benchmark datasets. The comparison results are listed in Tab.~\ref{tab:comparisons-of-three-streams}. We can observe that although only BLS is utilized compared with IPS, the MAE obviously decreases and F-measure scores increase significantly as shown in the second row. This indicates that the BLS provides a strong selectivity at object boundaries that boosts the performance a remarkable improvement. Besides, combined with TCS on the basis of the second setting, the performance of the model is further improved by amending the probable failures in transitional regions between boundaries and interiors. For example, the F$_{\beta}$ score increases from 0.914 to 0.929, with an improvement up to 1.5\% on HKU-IS dataset. We also provide examples of different settings. As shown in Fig.~\ref{fig:motivation-of-three-streams}, with the cooperation of IPS, BLS and TCS, the proposed method can generate more accurate results.

\textbf{Effectiveness of Integrated Successive Dilation Module.}
\begin{figure}[t]
\centering
\includegraphics[width=1\columnwidth,height=4.3cm]{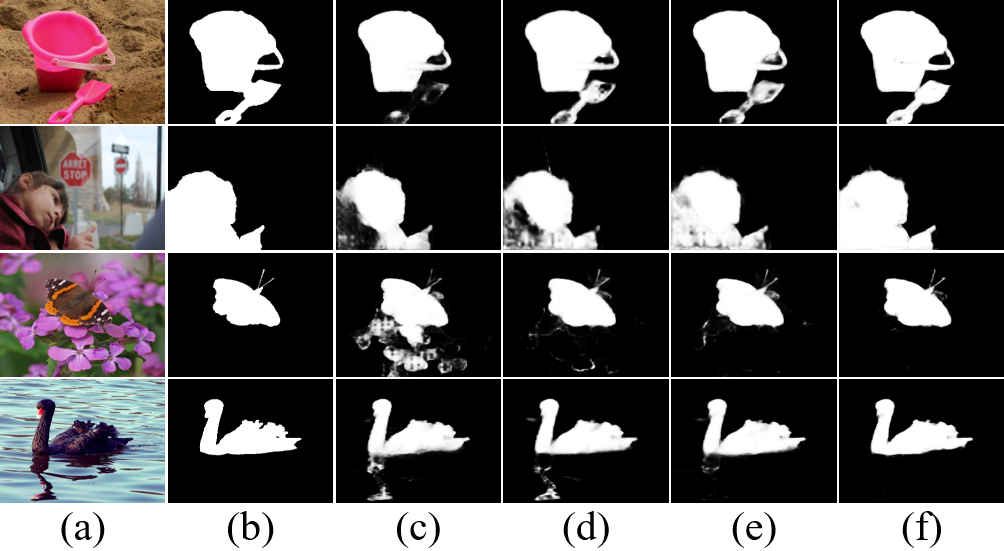}
\caption{Comparisons of ASPP and our ISD based on the proposed boundary-ware network. (a)~Images; (b)~ground-truth; (c)~without ISD; (d)~ASPP as a replacement of ISD; (e)~ASPP-M as a replacement of ISD; (f)~Our approach.}
\label{fig:result-of-modules}
\end{figure}
Atrous Spatial Pyramid Pooling (ASPP)~\cite{chen2018deeplab} is a common module for semantic segmentation~\cite{chen2018deeplab,chen2017rethinking,yang2018denseaspp}, which consists of multiple parallel convolutional layers with filters at different dilation rates of [6, 12, 18, 24] , thus capturing feature receptive fields at different scales.

To validate the effectiveness of our ISD, we construct three different models based on our BANet to compare on six benchmark datasets. The first network is that we remove ISD from our BANet. Secondly, we replace ISD with ASPP in BANet as the second network. Moreover, for a fairer comparison, we modify ASPP denoted as ``ASPP-M'' with the same branches and same dilation rates like ISD except for the short information flows and replace ISD with ASPP-M in our BANet as the third network.

The comparison of the three models and our BANet is listed in Tab.~\ref{tab:comparisons-of-different-modules}. From Tab.~\ref{tab:comparisons-of-different-modules}, we find that after ISD is removed from BANet, the performance of the method decreases dramatically on all the six datasets due to the lack of the capability to extract features for a region embedded in various contexts. As shown in the third row of Fig.~\ref{fig:result-of-modules}, the flowers close to the butterfly and the reflection of the goose in water are mistakenly detected as salient objects.

In fact, when the network utilizes ASPP or ASPP-M, the performance can also be improved to some extent. However, as the ISD has more information flow paths to aggregate the contextual information, the feature invariance can be enhanced better. Even if there are large appearance changes in the interiors of a large salient object, the whole salient object can be highlighted well, as shown in the last one line of Fig.~\ref{fig:result-of-modules}.

\section{Conclusion}
In this paper, we revisit the problem of SOD from the perspective of selectivity-invariance dilemma, where feature selectivity and invariance are required by different regions in salient objects. To solve this problem, we propose a novel boundary-aware network with successive dilation for salient object detection. In this network, boundary localization and interior perception streams are introduced to capture features with selectivity and invariance, respectively. A transition compensation stream is adopted to amend the probable failures between boundaries and interiors. Then the output of these three streams are fused to obtain the saliency mask in a boundary-aware feature mosaic selection manner. Moreover, we also propose a novel integrated successive dilation module for enhancing feature invariance to help perceiving and localizing salient objects. Extensive experiments on six benchmark datasets have validated the effectiveness of the proposed approach.

\section*{Acknowledgments}
This work was supported in part by the National Key R\&D Program of China under grant 2017YFB1002400, the National Natural Science Foundation of China (61672072, U1611461 and 61825101), and Beijing Nova Program under Grant Z181100006218063.

{\small
\bibliographystyle{ieee_fullname}
\bibliography{RefBANet}
}

\end{document}